\documentclass[lettersize,journal]{IEEEtran}
\usepackage{amsmath,amsfonts}
\usepackage{algorithmic}
\usepackage{algorithm}
\usepackage{array}
\usepackage[caption=false,font=normalsize,labelfont=sf,textfont=sf]{subfig}
\usepackage{textcomp}
\usepackage{stfloats}
\usepackage{url}
\usepackage{verbatim}
\usepackage{graphicx}
\usepackage{cite}
\usepackage{amsmath}
\usepackage{amssymb}
\usepackage{tablefootnote}
\usepackage{booktabs}
\usepackage[pagebackref,breaklinks,colorlinks]{hyperref}

\usepackage{multirow}
\usepackage{amsmath}
\usepackage{booktabs}
\usepackage[export]{adjustbox}
\usepackage{colortbl}

\newcommand{\cnn}{PillarNeSt}
\begin{document}

\title{PillarNeSt: Embracing Backbone Scaling and Pretraining for Pillar-based 3D Object Detection}

\author{Weixin Mao, Tiancai Wang, Diankun Zhang, Junjie Yan, Osamu Yoshie
\thanks{Weixin Mao and Tiancai Wang contributed equally to this work. Corresponding
author: Osamu Yoshie (e-mail: yoshie@waseda.jp).}}


\markboth{IEEE TRANSACTIONS ON INTELLIGENT VEHICLES}%
{Shell \MakeLowercase{\textit{et al.}}: A Sample Article Using IEEEtran.cls for IEEE Journals}


\maketitle

\begin{abstract}
This paper shows the effectiveness of 2D backbone scaling and pretraining for pillar-based 3D object detectors. Pillar-based methods mainly employ randomly initialized 2D convolution neural network (ConvNet) for feature extraction and fail to enjoy the benefits from the backbone scaling and pretraining in the image domain. To show the scaling-up capacity in point clouds, we introduce the dense ConvNet pretrained on large-scale image datasets (e.g., ImageNet) as the 2D backbone of pillar-based detectors. The ConvNets are adaptively designed based on the model size according to the specific features of point clouds, such as sparsity and irregularity. Equipped with the pretrained ConvNets, our proposed pillar-based detector, termed PillarNeSt, outperforms the existing 3D object detectors by a large margin on the nuScenes and Argoversev2 datasets. Our code shall be released upon acceptance.
\end{abstract}


\section{Introduction}\label{sec:intr}
\IEEEPARstart{P}{oint} cloud 3D object detection, a crucial task for autonomous driving and robotics, aims to identify and locate objects surrounding the ego agent. 
While great advance has been achieved in this field, there remains a significant gap in performance and efficiency, especially in the context of processing raw point cloud data. 
The primary approach to address this challenge is to transform point clouds into structured data, such as voxel-based and pillar-based representations. Voxel-based methods~\cite{SECOND, VoxelNet, centerpoint, largekernel3d2023} produce 3D voxels and employ 3D convolutions for feature extraction. 
Pillar-based~\cite{PointPillars, PillarNet, FastPillars, PillarNeXt} approaches first transform the point clouds into a pseudo-image representation and then employ a 2D backbone for feature extraction. 

Pillar-based methods do not rely on 3D  ConvNet to produce the features compared to voxel-based counterparts, achieving high detection speed and are relatively easier for practical deployment.
As an efficient paradigm, the pillar-based detectors mainly employ the randomly initialized 2D ConvNet for feature extraction. The architecture of 2D ConvNet is simply adopted from the one in image domain.
However, the 2D backbone~\cite{VGG15, ResNet16, EfficientNet19, ConvNeXt} pretrained on large-scale datasets (e.g., ImageNet~\cite{ImageNet09}) are not efficiently utilized for pillar-based detectors. Obvious scaling-up phenomenon with the model size is not observed for the point-cloud based detectors.
Meanwhile, the great progress has been observed in 2D image perception~\cite{ResNet16,ViT}. The 2D downstream perception tasks show large performance improvements with the backbone scaling and pretraining on large-scale image datasets. 
So we may ask two unresolved questions: (i) if it is possible to observe the scaling up phenomenon similar to the 2D image domain? (ii) how to transfer the image knowledge extracted by the pretraining process for the point cloud? For the first question, the biggest problem is the lack of reasonable and scalable backbone design for point cloud. While for the second problem, it is difficult to pretrain image backbones for adapting the special backbone design in pillar-based detector.

\begin{figure}[t]
	\includegraphics[width=0.99\columnwidth]{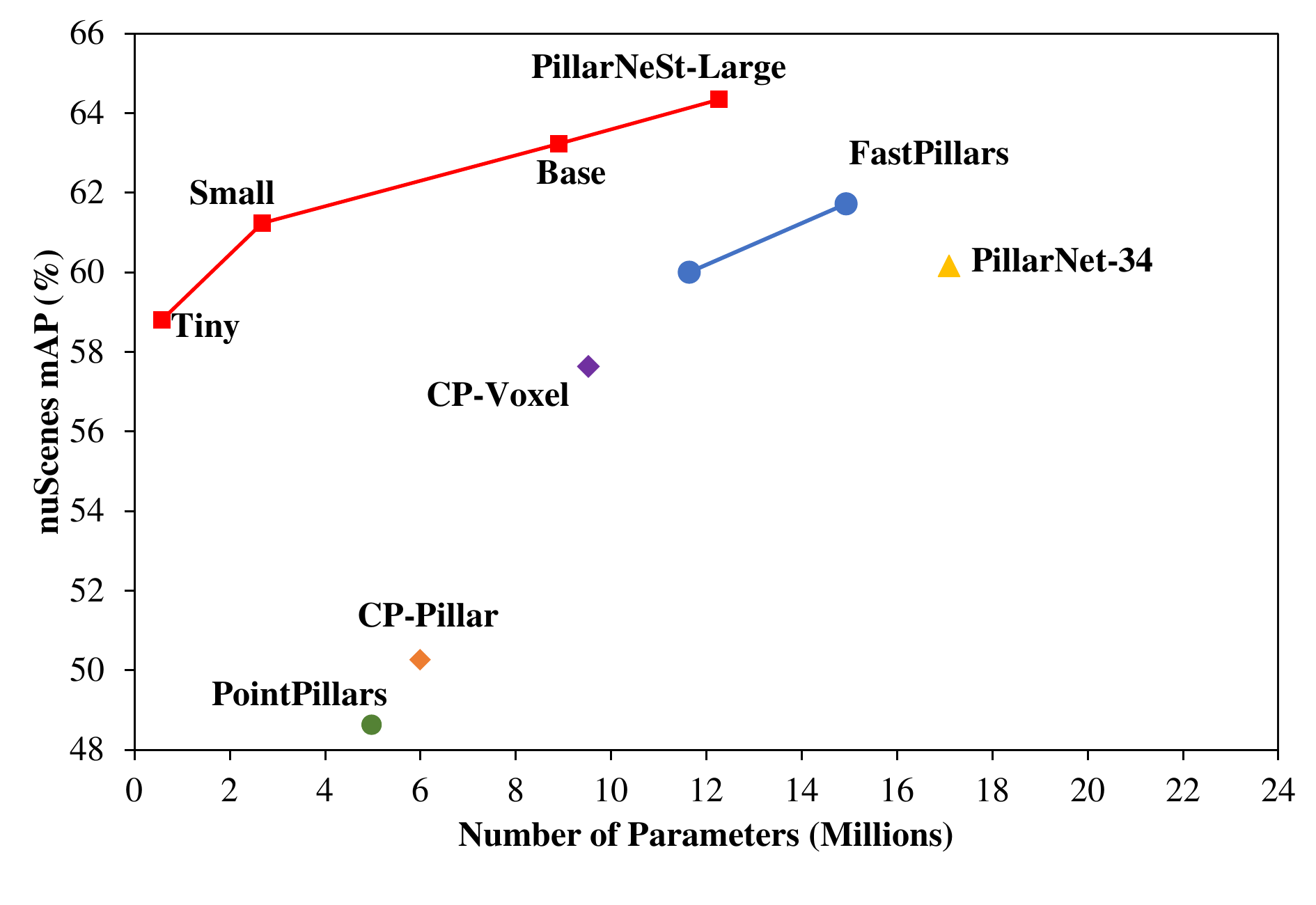}
        \hspace{-44mm}\resizebox{.44\columnwidth}{!}{
		\begin{tabular}[b]{l|cr}
			& mAP & \#Params  \\
                \hline
			CP-Pillar \cite{centerpoint} & 50.3\%  & 6.0M \\
			\bf \cnn{}-Tiny & \bf 58.8\%  & \bf 0.6M\\
			\hline
			CP-Voxel~\cite{centerpoint} & 57.6\% & 9.5M \\
			\bf \cnn{}-Small & \bf 61.7\%  & \bf 2.7M \\
			\hline
                PillarNet~\cite{PillarNet}  &  60.2\% & 17.1M \\
                \bf \cnn{}-Base & \bf 63.2\% & \bf8.9 M \\
			\hline
                FastPillar~\cite{FastPillars}   & 61.7\% & 11.6M \\
			\bf \cnn{}-Large & \bf 64.3\% & \bf 12.3M \\
                \vspace{16mm} \\ 
	\end{tabular}}
    \vskip -0.15in
	\caption{\textbf{Model Size vs. mAP.} All results are reported with a single model in nuScenes val set. CP: CenterPoint.}
	\label{fig:model_size}
\end{figure}

In this paper, we aim to answer these two questions above and explore the effectiveness of backbone \textit{scaling} and \textit{pretraining} for pillar-based 3D object detector. Since the pillar-based methods transform the point clouds into the pseudo-image representation, we mainly refer to the well-known ConvNeXt~\cite{ConvNeXt} in image domain as our basic network for dense feature extraction. 
To verify the effectiveness of backbone scaling, we first construct a strong baseline based on CenterPoint-Pillars\cite{centerpoint} with some modifications on the pillar encoder, detection head, and loss function.
Then, we modify the architecture of ConvNeXt to adapt the features of point cloud. In specific, we introduce several backbone design rules according to the characteristics of point cloud. For example, we introduce the large kernel (e.g., 7$\times$7 convolution layer) to increase the receptive field for point cloud and more blocks in early stages for refining fine-grained point cloud features.
After that, for the backbone scaling, we provide a series of ConvNet models with different scales, namely \cnn{}-Tiny, -Small, -Base, and -Large, to achieve the trade-off between the performance and inference speed. 
Finally, for the backbone pretraining, we utilize the ConvNeXts pretrained on large-scale image datasets (e.g., ImageNet~\cite{ImageNet09}) as the weight initialization for 2D backbone. Considering the large difference between the original ConvNeXt and our modified version, we introduce the weight initialization adaptation from two
perspectives: stage view and micro view.

With the special design, scaling and pretraining of 2D dense backbone, our proposed \cnn{} series show the scaling up of performance as the model size increases, as shown in Figure~\ref{fig:model_size}. 
Moreover, we observe that convergence is facilitated and performance is improved when using the pretrained 2D backbone, compared to the common random initialization. Our proposed PillarNeSt outperforms the existing 3D object detectors by a large margin on the nuScenes and Argoversev2 datasets.
Without any test-time augmentation, \cnn{}-L achieves \textbf{66.9\% mAP} and \textbf{71.6\% NDS} on nuScenes test set. \cnn{}-B achieves  \textbf{35.9\% mAP} and \textbf{28.6\% CDS} on Argoversev2 val dataset.
We hope the observations can bring some insights into integrating the scaling and pretraining 2D backbone for 3D object detection.

\section{Related Work}
\label{sec:rela_work}

\subsection{Grid-based 3D Object Detection}
Grid-based 3D object detection has gained significant interest in recent years, especially in applications like autonomous driving~\cite{wang2022performance, wang2023multi, chen2022milestones} and robotics. Grid-based split methods offer a structured way to handle the inherent sparsity and irregularity of point clouds. These methods are typically categorized into voxel-based and pillar-based approaches.

\noindent\textbf{Voxel-Based methods.} One of the earliest and most influential works in this domain is~\cite{VoxelNet}, which splits point clouds into 3D voxels and employs 3D convolutions for feature extraction. While VoxelNet sets the stage, it also introduces some challenges, such as empty voxels in outdoor environments. Subsequent methods, like~\cite{SECOND, VoxelRCNN21, centerpoint}, address these challenges by introducing 3D sparse convolutions, which not only enhance detection accuracy but also improve computation efficiency. ~\cite{largekernel3d2023, Link23} greatly improve the performance of voxel-based methods by optimizing the backbone.

\noindent\textbf{Pillar-Based methods.}
PointPillars~\cite{PointPillars} shifts the view from 3D voxelization to 2D pillar, focusing on the ground plane. By combining 2D voxelization with a Point-based feature extractor, PointPillars efficiently leverages 2D convolutions, making it particularly suitable for embedded systems with limited computational resources. 
Other works, such as~\cite{MVF17, HVNet20, HVPR21}, have further refined pillar-based detection by introducing feature projection and multi-scale aggregation strategies. 
Historically, pillar-based approaches trail voxel-based methods in terms of performance. Recently, ~\cite{PillarNet, FastPillars, PillarNeXt} introduce more advanced backbones, bridging the performance gap with voxel-based methods.

\subsection{Backbone Pretraining and Scaling}

Backbone pretraining and scaling~\cite{ResNet16, EfficientNet19, ViT, Swin} have shown great success in 2D perception tasks. By pretraining on large-scale datasets like ImageNet\cite{ImageNet09}, 2D ConvNets can capture general knowledge from images that can be transferred to downstream tasks. Model scaling, which improves the depth, width, and resolution of the network, has been a crucial strategy for improving the performance of various image recognition tasks. A notable example of the application of backbone scaling and pretraining is EfficientNet\cite{EfficientNet19}, which leverages a compound scaling method to uniformly scale all dimensions of the network.

In the 3D domain, methods like~\cite{VoxelMAE22, BEVMAE22}  utilize local reconstruction techniques for pretraining point cloud models. The pretraining transformer networks provide a robust initialization, which is beneficial for downstream tasks. 
~\cite{P2P22} utilizes pretrained image models for 3D analysis. By transferring knowledge from 2D image domains to 3D point clouds, P2P bridges the gap between these two modalities, potentially benefiting from the vast amount of labeled data available in 2D tasks.

\begin{figure*}[htbp]
\centering
\includegraphics[width=1.0\textwidth]{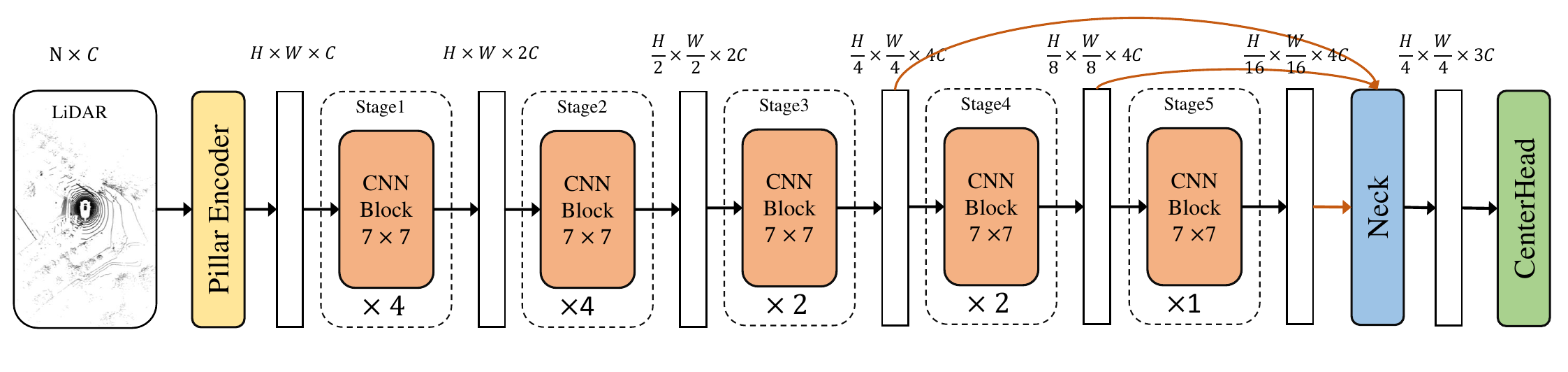}\vspace{-5mm}
\caption{\textbf{The overall architecture of a \cnn{} (\cnn{}-Base).} Our model takes raw point clouds as input and employs a pillar encoder, comprised of MLPs and max\&avg pooling, to transform the 3D data into a pseudo-image representation. It is then fed into a well-designed 2D dense ConvNet, pretrained on a large-scale image dataset. Multi-scale features of the last three stages are injected into the neck to produce fused features and Centerheads are applied for 3D box regression.}
\label{fig:arch}
\end{figure*}

\begin{table*}[t] 
\centering
\caption{\textbf{Detailed architectures for \cnn{} series backbones.}}
\label{tab:multi_bk}
\addtolength{\tabcolsep}{-2pt}
\vspace{4ex}
\scalebox{1.0}{
\begin{tabular}{c|c|c|c|c|c}
\hline
& \begin{tabular}[c]{@{}c@{}} output size\end{tabular} & Tiny & Small & Base & Large \\
\hline
\multirow{5}{*}{stage-1} & 
\multirow{5}{*}{\begin{tabular}[c]{@{}c@{}} 720$\times$720 \end{tabular}} & 
\multirow{5}{*}{$\begin{bmatrix}\text{d7$\times$7, 48}\\\text{1$\times$1, 192}\\\text{1$\times$1, 48}\end{bmatrix}$ $\times$ 2}  & 
\multirow{5}{*}{$\begin{bmatrix}\text{d7$\times$7, 48}\\\text{1$\times$1, 192}\\\text{1$\times$1, 48}\end{bmatrix}$ $\times$ 3} & 
\multirow{5}{*}{$\begin{bmatrix}\text{d7$\times$7, 64}\\\text{1$\times$1, 256}\\\text{1$\times$1, 64}\end{bmatrix}$ $\times$ 4} & 
\multirow{5}{*}{$\begin{bmatrix}\text{d7$\times$7, 96}\\\text{1$\times$1, 384}\\\text{1$\times$1, 96}\end{bmatrix}$ $\times$ 6} \\
& & & & \\
& & & & \\
& & & & \\
& & & & \\
\hline
\multirow{5}{*}{stage-2} & 
\multirow{5}{*}{\begin{tabular}[c]{@{}c@{}} 360$\times$360 \end{tabular}} & 
\multirow{5}{*}{$\begin{bmatrix}\text{d7$\times$7, 96}\\\text{1$\times$1, 384}\\\text{1$\times$1, 96}\end{bmatrix}$ $\times$ 2} &
\multirow{5}{*}{$\begin{bmatrix}\text{d7$\times$7, 192}\\\text{1$\times$1, 768}\\\text{1$\times$1, 192}\end{bmatrix}$ $\times$ 3} & 
\multirow{5}{*}{$\begin{bmatrix}\text{d7$\times$7, 192}\\\text{1$\times$1, 768}\\\text{1$\times$1, 192}\end{bmatrix}$ $\times$ 4} & 
\multirow{5}{*}{$\begin{bmatrix}\text{d7$\times$7, 192}\\\text{1$\times$1, 768}\\\text{1$\times$1, 192}\end{bmatrix}$ $\times$ 6} \\
& & & & \\
& & & & \\
& & & & \\
& & & & \\
\hline
\multirow{5}{*}{stage-3} & 
\multirow{5}{*}{\begin{tabular}[c]{@{}c@{}} 180$\times$180 \end{tabular}} & 
\multirow{5}{*}{$\begin{bmatrix}\text{d7$\times$7, 96}\\\text{1$\times$1, 384}\\\text{1$\times$1, 96}\end{bmatrix}$ $\times$ 1} & 
\multirow{5}{*}{$\begin{bmatrix}\text{d7$\times$7, 192}\\\text{1$\times$1, 768}\\\text{1$\times$1, 192}\end{bmatrix}$ $\times$ 2} & 
\multirow{5}{*}{$\begin{bmatrix}\text{d7$\times$7, 384}\\\text{1$\times$1, 1536}\\\text{1$\times$1, 384}\end{bmatrix}$ $\times$ 2} & 
\multirow{5}{*}{$\begin{bmatrix}\text{d7$\times$7, 384}\\\text{1$\times$1, 1536}\\\text{1$\times$1, 384}\end{bmatrix}$ $\times$ 4} \\
& & & & \\
& & & & \\
& & & & \\
& & & & \\
\hline
\multirow{5}{*}{stage-4} & 
\multirow{5}{*}{\begin{tabular}[c]{@{}c@{}} 90$\times$90 \end{tabular}} & 
\multirow{5}{*}{$\begin{bmatrix}\text{d7$\times$7, 96}\\\text{1$\times$1, 384}\\\text{1$\times$1, 96}\end{bmatrix}$ $\times$ 1} & 
\multirow{5}{*}{$\begin{bmatrix}\text{d7$\times$7, 192}\\\text{1$\times$1, 768}\\\text{1$\times$1, 192}\end{bmatrix}$ $\times$ 1} & 
\multirow{5}{*}{$\begin{bmatrix}\text{d7$\times$7, 384}\\\text{1$\times$1, 1536}\\\text{1$\times$1, 384}\end{bmatrix}$ $\times$ 2} & 
\multirow{5}{*}{$\begin{bmatrix}\text{d7$\times$7, 384}\\\text{1$\times$1, 1536}\\\text{1$\times$1, 384}\end{bmatrix}$ $\times$ 2} \\
& & & & \\
& & & & \\
& & & & \\
& & & & \\
\hline
\multirow{5}{*}{stage-5} & 
\multirow{5}{*}{\begin{tabular}[c]{@{}c@{}} 45$\times$45 \end{tabular}} & 
\multirow{5}{*}{$\begin{bmatrix}\text{d7$\times$7, 96}\\\text{1$\times$1, 384}\\\text{1$\times$1, 96}\end{bmatrix}$ $\times$ 1} & 
\multirow{5}{*}{$\begin{bmatrix}\text{d7$\times$7, 192}\\\text{1$\times$1, 768}\\\text{1$\times$1, 192}\end{bmatrix}$ $\times$ 1} & 
\multirow{5}{*}{$\begin{bmatrix}\text{d7$\times$7, 384}\\\text{1$\times$1, 1536}\\\text{1$\times$1, 384}\end{bmatrix}$ $\times$ 1} & 
\multirow{5}{*}{$\begin{bmatrix}\text{d7$\times$7, 384}\\\text{1$\times$1, 1536}\\\text{1$\times$1, 384}\end{bmatrix}$ $\times$ 2} \\
& & & & \\
& & & & \\
& & & & \\
& & & & \\
\hline
\multicolumn{2}{c|}{FLOPs (G)}
& 49 & 184 & 354 & 683 \\
\hline
\multicolumn{2}{c|}{\# params. (M)}
& 0.57 & 2.68 & 8.9 & 12.26 \\
\hline

\end{tabular}
}
\normalsize \vspace{-1mm}
\end{table*}

\section{Method}
\label{sec:mothod}

The architecture of our PillarNeSt is summarized in Figure~\ref{fig:arch}. Similar to the typical pillar-based detector, PillarNeSt comprises: (i) a pillar encoder to transform raw point clouds into the pseudo-image representation, (ii) a 2D backbone to extract features, (iii) a neck for fusing multi-scale features, and (v) a 3D detection head for predicting 3D objects. PillarNeSt mainly focuses on the 2D backbone part. We first introduce a strong baseline (Section~\ref{sec:StrongBaseline}) with some modifications on the pillar encoder, detection head, and loss function. Then we describe the backbone design (Section~\ref{sec:bk_design}), scaling (Section~\ref{sec:scaling}), and pretraining (Section~\ref{sec:pretrain}) strategies. Finally, we discuss the main differences with existing works (Section~\ref{sec:discussion}).

\subsection{Strong Baseline}\label{sec:StrongBaseline}
We choose CenterPoint-Pillars\cite{centerpoint} as our vanilla baseline, where PointPillars~\cite{PointPillars} is employed as the backbone. Our strong baseline mainly introduces some improvements on the pillar encoder, detection head and loss functions.

\noindent\textbf{Pillar encoder.} 
In the original Pillar encoder, Multi-layer Perceptions (MLPs) are used to extract features from point clouds, followed by max pooling to extract pillar features. 
Relying solely on max pooling leads to information loss. In this paper, we simultaneously employ max pooling and average pooling to preserve more information. Moreover, we also introduce the height offset of points, relative to the geometric center, as the input to compensate for the information loss on the $z$ axis.

\noindent\textbf{Detection head.} Previous works~\cite{AFDetV2, cia-ssd} have identified that a low correlation between classification and localization is one of the reasons for pool detection. Inspired by them, we integrate an IoU branch to predict the IoU score. 
During inference, we use IoU score to reweight classification score $\hat{S}$,
as illustrated in Eq.~\ref{eq:iou}:
\begin{equation}
    \hat{S} = \sqrt{S^{1-\alpha} * C^{\alpha}}
\label{eq:iou}
\end{equation}
Where $S$ is the classification score, $C$ is the predicted IoU score between $[0, 1]$, and $\alpha \in [0, 1]$ is a hyper-parameter to balance $S$ and $C$.

\noindent\textbf{Loss Functions.}
Following ~\cite{centerpoint}, we leverage the Gaussian Focal Loss for heatmap classification $\mathcal{L}_{cls}$, while employing the L1 loss to cater for regression offset $\mathcal{L}_{reg}$ and IoU score $\mathcal{L}_{iou}$. The composite loss $\mathcal{L}_{total}$ is formulated as:
\begin{equation}
     \mathcal{L}_{total} = \lambda_1\mathcal{L}_{cls} + \lambda_2\mathcal{L}_{iou} + \lambda_3\mathcal{L}_{reg}
\label{eq:loss}
\end{equation}
Here, $\lambda_1, \lambda_2, \lambda_3$ are corresponding weight coefficients.

\noindent\textbf{Training Strategy.}
During training, we use the copy-paste data augmentation from~\cite{centerpoint}. Following ~\cite{PointAugmenting21}, we introduce the fade strategy to disable the augmentation for the final several epochs.

\subsection{Backbone Design}\label{sec:bk_design}
\noindent\textbf{Large kernel matters.}
Some previous works~\cite{PillarNet, FastPillars, PillarNeXt} attempt to improve the representation by employing CNN with sparse or dense convolution layers. Those CNN backbones tend to employ $3\times3$ convolution layers for feature extraction. As mentioned above, the pillar-based methods apply the pillar encoder to transform the point cloud into 2D representations (pseudo image). Processing such 2D representation is quite similar to the case in the image domain. In the seminal work of ~\cite{ERF}, the concept of the Effective Receptive Field (ERF) is introduced. It points out that the ERF is not significantly improved by increasing the number of layers. Some recent works~\cite{RepLKNet,ConvNeXt} argue that large ERF can be effectively achieved by employing convolution with larger kernels. Moreover, ~\cite{SST, largekernel3d2023, Link23} also points out that a larger receptive field contributes to enhancing the capability of point cloud detectors. To balance efficiency and performance, we adopt ConvNeXt Block~\cite{ConvNeXt} with 7$\times$7 depth-wise convolution layer.

\noindent\textbf{Removing downsampling in the first stage.} Considering the information redundancy of images, the image backbone usually performs the downsampling operations, e.g., convolution layer with stride 2, to reduce the overall computation cost. Raw point cloud data is sparse and irregular, while containing rich detailed information. These features pose a challenge: premature or excessive down-sampling may lead to the loss of critical information~\cite{FCOSLiDAR22, PillarNet}. It underscores the importance of designing strategies that meticulously maintain these details while efficiently handling point clouds. 
Given these considerations, our backbone design removes the stem and refrains from implementing down-sampling in the first stage block. This strategic choice ensures the preservation of the original resolution of the input features. 

\noindent\textbf{More blocks in early stage.}
Backbone design in image domain~\cite{VGG15, ResNet16, ConvNeXt, ViT} tends to employ more block layers in later stages to extract high-level features for classification. For example, ResNet-101~\cite{ResNet16} for image classification takes 23 residual blocks in the 3rd stage while the 1st stage uses 3 blocks for feature extraction.
Considering that the point clouds are sparse and irregular, it means that only after multiple convolutions can make empty pillar features filled by the features diffused from occupied pillars. Different from the practice in the image domain, more convolution layers are supposed to be stacked in the early stages. In our practice, we stack more blocks in the early stages for fully refining fine-grained point cloud features. 
Our extensive experiments suggest that increasing the number of blocks in early stages yields superior gains compared to adding more blocks in later stages.

\noindent\textbf{More depth stage.} Through our analysis of point cloud scenarios, we observed a large variation in object sizes. For example, when the pillar size is set to 0.2m, the maximum perceivable range is 1.6m after $8\times$ downsampling. However, many objects in real-world scenarios exceed the limited perceivable range. It means the feature points after $8\times$ down-sampling cannot fully perceive the entire object for large objects. To tackle this issue, PillarNeXt~\cite{PillarNeXt} employs a complex ASPP~\cite{ASPP17} module to achieve large receptive fields. Different from it, we adopt a simple way and add one more stage (named stage-5) on top of stage-4, which contains only one or two ConNeXt blocks. The block number of stage-5 can be scaled up based on the model size. The output of added stage-5 is served as one of the multi-scale inputs of the neck network.  

\subsection{Backbone Scaling}\label{sec:scaling}
In this section, we design a series of 2D backbones, ranging from \cnn{}-Tiny with light weight to \cnn{}-Large with high performance, for different demands.
Building upon the design principles in Section~\ref{sec:bk_design}, we propose our \cnn{}-Tiny, Small, Base and Large models. As shown in Table~\ref{tab:multi_bk}, all model versions share a similar architecture. Each model consists of five stages, with stage-1 keeping the feature map size without downsampling. For stage-2\_1, stage-3\_1, stage-4\_1, and stage-5\_1, the features are downsampled by a factor of two. 
Each ConvNext block contains a depth-wise convolution layer with a kernel size of 7, followed by two $1\times1$ convolution layers. The architecture hyper-parameters of these model variants are:
\begin{itemize}
  \item \cnn{}-T: $C_{in}$ = 48, block numbers = $\{2, 2, 1, 1, 1\}$
  \item \cnn{}-S: $C_{in}$ = 48, block numbers = $\{3, 3, 2, 1, 1\}$
  \item \cnn{}-B: $C_{in}$ = 64, block numbers = $\{4, 4, 2, 2, 1\}$
  \item \cnn{}-L: $C_{in}$ = 96, block numbers = $\{6, 6, 4, 2, 2\}$
\end{itemize}
where $C_{in}$ is the input channel number of the first stage. The model size and theoretical computational complexity (FLOPs) are listed in Table~\ref{tab:multi_bk}.

\subsection{Backbone Pre-training}\label{sec:pretrain} 
\cnn{} aims to integrate the knowledge from backbone pretraining on large-scale image datasets, e.g., ImageNet~\cite{ImageNet09}. Since our backbone design is conducted based on ConvNeXt~\cite{ConvNeXt} architecture, we wonder if it is possible to utilize the ConvNeXt pre-trained on ImageNet for initialization. However, as mentioned above, the block number of each stage or the channel numbers in \cnn{} are different from the original ConvNeXt design. To resolve this problem, we introduce the weight initialization adaptation from two perspectives: stage view and micro view. (i) From the stage view, we simply copy the weights from pretrained ConvNeXt model for stage-1$\sim$4 while our added last stage (stage-5) is randomly initialized. For stage-1$\sim$4, if the block number is less than the ConvNeXt one, we only copy the parameters of corresponding blocks according to the block identity. (ii) From the micro view, we replicate the parameters from pretrained model for the first $C_{in}$ channels (or blocks) and the parameters of the left channels (or blocks)  are randomly initialized. 

\subsection{Discussion}
\label{sec:discussion}
Our work is somewhat similar to the PillarNet~\cite{PillarNet}, FastPillars~\cite{FastPillars} and PillarNeXt~\cite{PillarNeXt} from the perspective of backbone scaling. However, there are many differences between our PillarNeSt and them. PillarNet simply employs the randomly initialized VGG~\cite{VGG15}, ResNet-18~\cite{ResNet16} and ResNet-34 as the 2D backbone. It fails to scale the backbone from a unified architecture and enjoy the benefit from the image pretraining. FastPillars use a structural re-parameterization technique~\cite{RepVGG} to VGG and ResNet-34 for fast inference. PillarNeXt is the most close work to ours. However, it does not exhaustively design the backbones according to the features of point cloud, without the large kernel and more blocks in early stage. Our PillarNeSt aims to explore the effectiveness of both the backbone scaling and pretraining for the pillar-based 3D object detection. With specific design for point cloud, it shows that the detection performance is improved with the scaling-up of 2D backbone. Moreover, it is the first one to show that the point cloud based detectors can directly benefit from the 2D image pretraining.

\begin{table*}[t]
\centering
\caption{\textbf{A comparison of \cnn{} models of different scales with state-of-the-art methods on nuScenes validation set.} 'P', 'V', and 'R' represent pillar, voxel, and range-view-based grid encoders, respectively. ${\dagger}$: reported in~\cite{FastPillars}.} 
\label{tab:nus_val}
\begin{tabular}{l|c|c|c|c|ccccc}
\specialrule{1pt}{0pt}{1pt}
\toprule
Method & Encoder & Grid Size & NDS & mAP & mATE$\downarrow$ & mASE$\downarrow$ & mAOE$\downarrow$ & mAVE$\downarrow$ & mAAE$\downarrow$   \\ 
\hline
 CenterPoint~\cite{centerpoint}  & V &  0.075  &  66.8 & 59.6 & 29.2 & 25.5 & 30.2 & 25.9 & 19.3 \\
 HotSpotNet~\cite{HotSpotNet2020} & V & 0.1 & 66.0 & 59.5  &  -  &  -  &  -  &  -  &  - \\
Transfusion-L~\cite{TransFusion22} & V  &  0.075 & 66.8 & 60.0  &  -  &  -  &  -  &  -  &  -    \\ 
UVTR-L~\cite{UVTR2022} & V  &  0.075  & 67.7  & 60.9  & 33.4  & 25.7   & 30.0 & 20.4  & 18.2 \\  
VISTA~\cite{VISTA22} & V+R & 0.1 & 68.1 & 60.8  &  -  &  -  &  -  &  -  &  -    \\ 
LargeKernel3D~\cite{largekernel3d2023} & V & 0.075& 69.1 & 63.9 & -  &  -  &  -  &  -  &  - \\
LinK~\cite{Link23} & V & 0.05 & 69.5 & 63.6 & -  &  -  &  -  &  -  &  - \\
\hline
CenterPoint-Pillars~\cite{centerpoint} & P & 0.2 & 59.5 & 49.2 & 32.1 & 26.0 & 39.1 & 35.3 & 18.9 \\
PillarNet-18~\cite{PillarNet} & P &  0.075 & 67.4 & 59.9 & - & - & - & - & - \\ 
PillarNet-34${\dagger}$~\cite{PillarNet} & P & 0.075 & 67.6  & 60.2 & - & - & - & - & - \\ 
FastPillars-$m$~\cite{FastPillars} & P  &  0.15 & 68.2 & 61.7 & -  &  -  &  -  &  -  &  - \\
PillarNeXt-B~\cite{PillarNet}  & P & 0.075 & 68.8 & 62.5  &  27.8 & 25.1 & 26.9 & 24.8 & 20.1\\
\hline
\rowcolor[gray]{.9} \cnn{}-Tiny (Ours) & P & 0.15 & 65.6 & 58.8 & 29.2 & 25.1 & 31.2 & 33.1	& 19.0 \\
\rowcolor[gray]{.9} \cnn{}-Small (Ours) & P & 0.15 & 68.1 & 61.7 & 27.7 & 25.0 & 26.8 & 27.9	& 20.1 \\
\rowcolor[gray]{.9} \cnn{}-Base (Ours) & P & 0.15 & 69.2 & 63.2 & 26.6	& 24.6 & 26.8 & 27.0 & 19.3 \\
\rowcolor[gray]{.9} \cnn{}-Large (Ours) & P & 0.15 & \textbf{70.4} & \textbf{64.3} & 26.6 & 24.6 & 23.7 & 23.4 & 19.1 \\
\bottomrule
\specialrule{1pt}{1pt}{0pt}
\end{tabular}
\vspace{-2mm}
\vspace{-1mm}

\end{table*}

\begin{table*}[htbp]
\centering 
\caption{\textbf{State-of-the-art comparisons on nuScenes test set.} We show NDS, and mAP for each class. Abbreviations: construction vehicle (CV), pedestrian (Ped), motorcycle (Motor), bicycle (BC)  and traffic cone (TC). All models do not use any TTA or model ensembling.}  
 \label{tab:nusc_test}
\begin{tabular}{l|c|c|cccccccccc}
  \specialrule{1pt}{0pt}{1pt}
  \toprule 
  Method & NDS(\%) &mAP(\%) & Car & Truck & Bus & Trailer & CV & Ped & Motor & BC & TC & Barrier \\ 
 \midrule
 PointPillars \cite{PointPillars} & 45.3 & 30.5 & 68.4 & 23.0 & 28.2 & 23.4 & 4.1 & 59.7 & 27.4 & 1.1 & 30.8 & 38.9\\ 
 3DVID \cite{3DVID20} & 53.1 & 45.4 & 79.7 & 33.6 & 47.1 & 43.1 & 18.1 & 76.5 & 40.7 & 7.9 & 58.8 & 48.8\\ 
 3DSSD \cite{3DSSD20} & 56.4 &42.6  & 81.2 & 47.2 & 61.4 & 30.5 & 12.6 & 70.2 & 36.0 & 8.6 & 31.1 & 47.9\\ 
  Cylinder3D \cite{Cylinder3D20} & 61.6 & 50.6 & - & - & - & - & - & - & - & - & - & - \\
  CBGS \cite{CBGS19} & 63.3 & 52.8 & 81.1 & 48.5 & 54.9 & 42.9 & 10.5 & 80.1 & 51.5 & 22.3 & 70.9 & 65.7\\ 
  Pointformer \cite{Pointformer21} & - & 53.6 & 82.3 & 48.1 & 55.6 & 43.4 & 8.6 & 81.8 & 55.0 & 22.7 & 72.2 & 66.0\\
  CVCNet \cite{CVCNet20} & 64.4 & 55.3 & 82.7 & 46.1 & 46.6 & 49.4 & 22.6 & 79.8 & 59.1 & 31.4 & 65.6 & 69.6 \\ 
  Centerpoint \cite{centerpoint} & 65.5 & 58.0 & 84.6 & 51.0 & 60.2 & 53.2 & 17.5 & 83.4 & 53.7 & 28.7 & 76.7 & 70.9 \\ 
  FCOS-LIDAR \cite{FCOSLiDAR22} & 65.7 & 60.2 & 82.2 & 47.7 & 52.9 & 48.8 & 28.8 & 84.5 & 68.0 & 39.0 & 79.2 & 70.7 \\ 
  HotSpotNet \cite{HotSpotNet2020} & 66.0 & 59.3 & 83.1 & 50.9 & 56.4 & 53.3 & 23.0 & 81.3 & 63.5 & 36.6 & 73.0 & 71.6\\ 
  VMVF~\cite{VMVF22} & 67.3 & 60.9 & 84.6 & 50.0 & 63.2 & 55.3 & 23.4 & 83.7 & 65.1 & 38.9 & 76.8 & 68.2\\
  AFDetV2 \cite{AFDetV2} & 68.5 & 62.4 & 86.3 & 54.2 & 62.5 & 58.9 & 26.7 & 85.8 & 63.8 & 34.3 & 80.1 & 71.0\\
  VISTA \cite{VISTA22} & 69.8 & 63.0 & 84.4 & 55.1 & 63.7 & 54.2 & 25.1 & 82.8 & 70.0 & 45.4 & 78.5 & 71.4 \\ 
  TransFusion-L~\cite{TransFusion22} & 70.2 & 65.5 & 86.2 & 56.7 & 66.3 & 58.8 & 28.2 & 86.1 & 68.3 & 44.2 & 82.0 & 78.2 \\
  LargeKernel ~\cite{largekernel3d2023}& 70.5 & 65.3 & 85.9 & 55.3 & \textbf{66.2} & 60.2 & 26.8 & 85.6 & 72.5 & 46.6 & 80.0 & 74.3 \\
  LinK ~\cite{Link23}& 71.0 & 66.3 & 86.1 & 55.7 & 65.7 & 62.1 & 30.9 & 85.8 & \textbf{73.5} & 47.5 & 80.4 & 75.5 \\
  \rowcolor[gray]{.9} \cnn{}-Base (Ours) &  71.3 & 65.6 & 87.1 & 55.5 & 61.6 & 62.1 & \textbf{31.0} & 86.3 & 69.4 & 46.8 & 80.6 & 76.0\\
  \rowcolor[gray]{.9} \cnn{}-Large (Ours) & \textbf{71.6} & \textbf{66.9} & \textbf{87.4} & \textbf{56.4} & 64.0 & \textbf{63.0} & 30.7 & \textbf{86.6} & 69.4 & \textbf{51.5} & \textbf{82.3} & \textbf{77.9}\\
  \bottomrule
  \specialrule{1pt}{1pt}{0pt}
 \end{tabular}\vspace{-2mm}
\vspace{-0.5em}
\end{table*}

 
\begin{table*}[ht]
\caption{
		\textbf{Comparison with state-of-the-art methods on Argoverse2 validation split.}
		\dag: provided by authors of AV2 dataset.
		\ddag: reimplemented by us.
		$^\ast$: reimplemented by FSD \cite{FSD22}. \S: the proposed method with smaller pillars.
            The average results consider all categories.
	}
    \label{tab:sota_argo}
	\begin{center}
		\resizebox{\textwidth}{!}{%
			\begin{tabular}{l|l|ccccccccccccccccccccccccc}
				\specialrule{1pt}{0pt}{1pt}
				\toprule
				&
				\textbf{Methods} &
				\rotatebox{90} {Average}& 
				\rotatebox{90} {Vehicle} & 
				\rotatebox{90} {Bus} &
				\rotatebox{90} {Pedestrian} &
				\rotatebox{90} {Box Truck} &
				\rotatebox{90} {C-Barrel} &
				\rotatebox{90} {Motorcyclist} &
				\rotatebox{90} {MPC-Sign} &
				\rotatebox{90} {Motorcycle} &
				\rotatebox{90} {Bicycle} &
				\rotatebox{90} {A-Bus} &
				\rotatebox{90} {School Bus} &
				\rotatebox{90} {Truck Cab} &
				\rotatebox{90} {C-Cone} &
				\rotatebox{90} {V-Trailer} &
				\rotatebox{90} {Bollard} &
				\rotatebox{90} {Sign} &
				\rotatebox{90} {Large Vehicle} &
				\rotatebox{90} {Stop Sign} &
				\rotatebox{90} {Stroller} &
				\rotatebox{90} {Bicyclist} &
				\\ 
				\midrule
				\multirow{6}{*}{mAP} &
				CenterPoint\dag \cite{centerpoint}   & 13.5 & 61.0 & 36.0  & 33.0  & 26.0    & 22.5  & 16.0   & 16.0 & 12.5 & 9.5  & 8.5 & 7.5 & 8.0 & 8.0  & 7.0 & 25.0 & 6.5  &3.0 & 28.0 & 2.0 & 14  \\
				&CenterPoint$^\ast$ \cite{centerpoint}        & 22.0 & 67.6 & 38.9 & 46.5 & 40.1 & 32.2 & 28.6 & 27.4 & 33.4 & 24.5 & 8.7 & 25.8 & 22.6 & 29.5  & 22.4 & 37.4 & 6.3 & 3.9 & 16.9 & 0.5 & 20.1\\
				&CenterPoint\ddag  \cite{centerpoint}     & 27.4	&78.7&	45.3&	62.2&	43.7&	59.3&	10.3&	45.4&	33.2&	30.6&	25.8&	27.2&	23.2&	36.7&	23.8&	50.3&	15.0&	4.9&	47.9&	0.2	& 15.2\\
				&FSD \cite{FSD22}     & 28.2 & 68.1 & 40.9 & 59.0 & 38.5 & 42.6 & 39.7 & 26.2 & 49.0 & 38.6 & 20.4 & 30.5 & 14.8 & 41.2  & 26.9 & 41.8 & 11.9 & 5.9 & 29.0 & 13.8 & 33.4\\
				&VoxelNeXt \cite{VoxelNeXt23}     & 30.5 & 72.0 & 39.7 & 63.2 & 39.7 & 64.5 & 46.0 & 34.8 & 44.9 & 40.7 & 21.0 & 27.0 & 18.4 & 44.5  & 22.2 & 53.7 & 15.6 & 7.3 & 40.1 & 11.1 & 34.9\\
				\rowcolor[gray]{.9} &\cnn{}-Base (Ours) &32.7 &	78.4&	44.3& 	65.9 &	42.0 &	57.0&	32.3 &	61.0 &	46.4 	&39.9 &	28.1 &	39.1& 	25.1& 	51.7 &	28.7 &	59.3 &	19.7 &	4.6 &	48.5 &	6.0 &	20.0 \\
				\rowcolor[gray]{.9} &\cnn{}-Base\S (Ours) & \textbf{35.9} & 78.6 &	47.0& 	72.0& 	45.0& 	68.6 &	30.1 &	54.2 &	53.2& 	44.8 &	26.1 &	38.1 & 24.2 &	56.2 &	29.5 &	66.3 &	21.9 &	3.9 &	47.7 &	18.0 &	26.9\\ 
				\midrule
				\multirow{5}{*}{CDS} &
				CenterPoint$^\ast$ \cite{centerpoint}    & 17.6  & 57.2  & 32.0  & 35.7   & 31.0   & 25.6  & 22.2 & 19.1 & 28.2  & 19.6 & 6.8  & 22.5 & 17.4  & 22.4 & 17.2 & 28.9 & 4.8 & 3.0 & 13.2 & 0.4 & 16.7  \\
				&CenterPoint\ddag \cite{centerpoint}     & 21.0&	66.5&	37.1&	45.1&	35.6&	45.5&	7.0&	31.0&	24.0&	22.7&	20.8&	22.6&	17.8&	26.2&	17.4&	35.4&	11.1&	3.5&	37.8&	0.2	&11.6 \\
				&FSD \cite{FSD22}         & 22.7 & 57.7   & 34.2  & 47.5 & 31.7   & 34.4  & 32.3 & 18.0 & 41.4  & 32.0 & 15.9 & 26.1 & 11.0  & 30.7 & 20.5 & 30.9 & 9.5 & 4.4 & 23.4 & 11.5 & 28.0 \\
				&VoxelNeXt \cite{VoxelNeXt23}                & 23.0 & 57.7   & 30.3  & 45.5 & 31.6   & 50.5  & 33.8 & 25.1 & 34.3  & 30.5
				& 15.5 & 22.2 & 13.6  & 32.5 & 15.1 & 38.4 & 11.8 & 5.2 & 30.0 & 8.9 & 25.7 \\
				\rowcolor[gray]{.9} &\cnn{}-Base (Ours) &25.7 &	67.6 &	37.0 &	49.1 &	34.7& 	44.1 &	25.5 &	42.6& 	37.8& 	31.0& 	22.9 &	33.7 &	19.7 &	38.4 &	21.6 &	43.1 &	15.7 &	3.3 &	38.8 &	4.6 &	16.1 \\
				\rowcolor[gray]{.9} &\cnn{}-Base\S (Ours) & \textbf{28.6} &	67.7 &	39.7 &	55.7 &	37.3 &	55.2 &	23.8 &	37.0 &	44.5 &	36.3 &	21.3 &	33.2 &	19.0 &	42.1 &	23.0 &	48.5 &	17.7 &	2.8 &	38.8 &	14.6 &	21.5\\ 

				\bottomrule
				\specialrule{1pt}{1pt}{0pt}
			\end{tabular}%
   }
	\end{center}\vspace{-3mm}
\end{table*}

\section{Experiment}
\label{sec:exp}


\noindent\textbf{nuScenes Dataset.} nuScenes~\cite{nuscenes} is a sizable and frequently used dataset. It consists of multi-modal data obtained from 1000 scenes, including RGB images from six surround-view cameras, radar points from five different radars, and point cloud data from one LiDAR. It is broken down into 700/150/150 scenarios for training/validation/testing. There are eleven categories and 1.4 million annotated 3D bounding boxes in total. 
We assess the performance using the mean Average Precision (mAP) and the nuScenes Detection Score (NDS). The NDS offers a comprehensive measure by encompassing various aspects of detection capabilities.

\noindent\textbf{Argoverse2 Dataset.} 
Argoverse2 (AV2)~\cite{Argoversev2} is a large-scale dataset for long-range perception and it contains 1000 scenes in total, 700 for training, 150 for validation, and 150 for testing. AV2 also adopts a composite metric called Composite Detection Score (CDS) as an addition to the commonly used mAP. Similar to the NDS of nuScenes, CDS takes both mAP and other localization errors into account. The perception range in AV2 reaches 200 meters (total area $400m \times 400m$), which is much larger than nuScenes. We conduct experiments on AV2 dataset to validate the long-range performance of the proposed method.

\noindent\textbf{Implementation Details.} \cnn{} models are trained by using Adam optimizer and the one-cycle learning rate schedule using the mmdetection3d~\cite{mmdet3d2020} framework. All ablation experiments are performed on 8 V100 GPUs, while other experiments are conducted on 8 A100 GPUs. $\lambda_1, \lambda_2, \lambda_3$ are set to 1.0, 1.0 and 0.25, respectively. Furthermore, during training, we also use the ground-truth copy-paste data augmentation from~\cite{centerpoint} and disable this augmentation for the final 5 epochs, following ~\cite{PointAugmenting21}.

For nuScenes, the one-cycle learning rate policy with an initial learning rate of 1e-3 is used. we use an AdamW optimizer. The weight decay is 0.01, and the momentum ranges from 0.85 to 0.95. In the horizontal plane, we set the point cloud range to [-54m, 54m], and in the vertical direction, we set the point cloud range to [-5m, 3m]. For PillarNeSt, we set the pillar size as [0.15m, 0.15m, 8m]. Moreover, following CenterPoint, we employ class-agnostic NMS as the post-processing during inference, with the score threshold set to 0.2 and the rectification factor $\alpha$ set to 0.5 for the ten classes. 
For AV2, we use an Adam optimizer, with 0.01 weight decay and a base learning rate of 1e-4. The point cloud range is set to [-153.6m, 153.6m] for both the X-axis and Y-axis, and [-5m, 5m] for the Z-axis. The post-processing is the same as nuScenes. Note that the results are evaluated in the range [-200m, 200m] in the horizontal plane.

\begin{table} [htbp] 
\centering
\caption{\textbf{Strong baseline on the nuScenes val set.} $\dag$: reproduced results based on the official codebase. }
\label{tab:strong_baseline}
\begin{tabular}{l|ccc}
\toprule
Methods & mAP$\uparrow$ & mAOE$\downarrow$ & NDS$\uparrow$\\
\midrule
CenterPoint-Pillar $\dag$ & 49.7 & 37.0 & 60.1 \\
\midrule
$+$ Strong baseline & \textbf{54.8} & \textbf{29.6} & \textbf{63.5} \\
\bottomrule
\end{tabular}\vspace{-2mm}
\end{table}

\begin{table}[t]
\centering
\caption{ \textbf{Ablation study of adding more depth stage.} The second row adds one more stage with $2\times$ downsampling. 
}
\label{tab:stage5}
  \begin{tabular}{c|c|c|c}
\toprule
Backbone Stages & FLOPs (G) & mAP & NDS \\
\midrule
$(1\times, 2\times, 4\times, 8\times)$ & 37.06 & 49.8 & 59.9  \\
$(1\times, 2\times, 4\times, 8\times, 16\times)$ & 37.52 & \textbf{51.8} & \textbf{61.2} \\
\bottomrule
  \end{tabular}\vspace{-2mm}
\end{table}

\subsection{ Overall Results}

\subsubsection{Strong Baseline.}
In Section~\ref{sec:StrongBaseline}, we improve the CenterPoint-Pillar baseline to a strong version. As shown in Table~\ref{tab:strong_baseline}, our strong baseline achieves a 54.5\% mAP and 63.5\% NDS, compared to the original baseline with 49.7\% mAP and 60.1\% NDS. It should be noted that the great improvements on mAOE are mainly from the introduced IoU branch in CenterPoint head while the fade strategy contributes a lot to improve the mAP and NDS metrics.

\subsubsection{Comparison with State-of-the-Art on nuScenes}
For a fair comparison, we evaluate all our \cnn{} models against previously published LiDAR-only methods on the nuScenes validate/test set. 
As shown in Table~\ref{tab:nus_val}, our \cnn{}-Large model achieves to \textbf{70.4\% NDS} and \textbf{64.3\% mAP}, outperforming all previous methods. Specifically, our \cnn{}-Large model surprisingly surpasses the previously advanced Pillar-based method, PillarNeXt-B, by +1.6\% in NDS and +1.8\% in mAP. 
Notably, our model exceeds all voxel-based methods, such as LargeKernel3D~\cite{largekernel3d2023} and LinK~\cite{Link23}.
We also showcase our results on the nuScenes test. As shown in Table~\ref{tab:nusc_test}, our approach still achieves state-of-the-art (SOTA) performance compared to all point cloud 3D object detection methods. Specifically, our Base model achieves an NDS of 71.3\% and an mAP of 65.6\%, while our Large model achieves \textbf{71.6\% NDS} and \textbf{66.9\% mAP}.

\subsubsection{Comparison with State-of-the-Art on Argoverse2}
We also evaluate the proposed method on the \textit{val} set of long-range Argoverse2 dataset in Table \ref{tab:sota_argo}. Due to the huge cost caused by the long range, we only report results using \cnn{}-Base backbone. \cnn{}-Base outperforms the existing LiDAR-based detection methods ~\cite{VoxelNeXt23} by \textbf{+5.4 mAP} and \textbf{+5.6 CDS} on the average metrics. 
The per-class evaluation results are also listed, which shows that \cnn{} has a great advantage in detecting larger objects (i.e. Vehicle, Bus, V-Trailer, etc.). The improvements are benefited from the large receptive field of \cnn{}.

\subsection{Ablation Study}
In this section, we explore the effectiveness of our backbone design principles and the scaling rules on nuScenes val set. For our baseline, the block number of four stages is set to [1, 1, 1, 1] and the channel number of the first stage is 48.

\begin{figure}[t]
\centering
\includegraphics[width=0.5\textwidth]{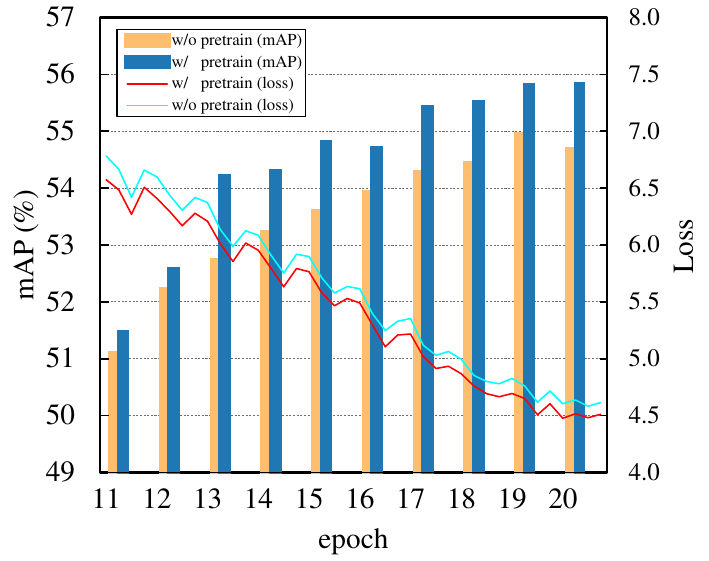} 
\vspace{-6mm}
\caption{\textbf{Ablation study of pretraining.} Pretraining initialization achieves faster convergence and higher mAP.}
\label{fig:pretrain}
\end{figure}

\begin{figure*}[htbp]
\centering
\includegraphics[width=1.0\textwidth]{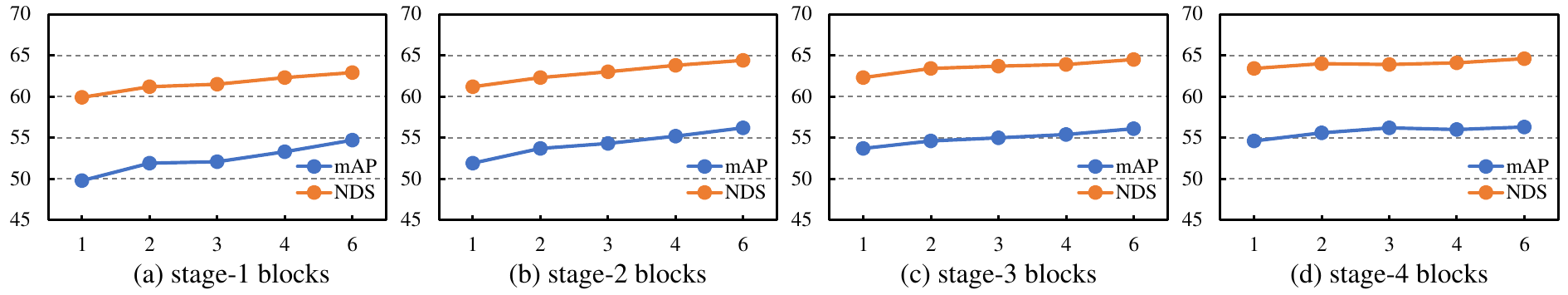}\vspace{-2mm}
\caption{\textbf{Impact of varying block quantities in the four stages of the backbone on the final results.} Each subfigure corresponds to a different stage of the backbone. Subfigure $(a)\sim(d)$ shows the effect of changing the number of blocks in Stage $1\sim4$, respectively. As can be seen, different stages show varying sensitivity to the number of blocks utilized.}
\label{fig:num_blocks}
\end{figure*}

\begin{table}[htbp]
\centering
\caption{ \textbf{Ablation study of stage-1 downsampling.} The experiment in second row uses a downsampling in stage-1. 
}
\label{tab:stage1_down}
  \begin{tabular}{c|c|c|c}
\toprule
Backbone Stages & mAP & NDS & FLOPs (G) \\ 
\midrule
$(1\times, 2\times, 4\times, 8\times)$ & \textbf{49.8} & \textbf{59.9} & 37 \\
$(2\times, 4\times, 8\times, 16\times)$ & 42.8 & 54.6 &  10 \\

\bottomrule
  \end{tabular}\vspace{-2mm}
\end{table}

\begin{table}[t]
  \centering
  \caption{ \textbf{Ablation study for varying input channels of the backbone.} "input\_ch" means input channels. 
  }
  \label{tab:ab_input}
  \begin{tabular}{c|cc|c|c}
    \toprule
input\_ch & mAP & NDS & FLOPs (G) & \# params. (M) \\
\hline
32 & 49.2 & 59.4 & 33 & 1.25 \\
48 & 49.8 & 59.9 & 37 & 1.27 \\
64 & 50.7 & 60.5 & 42 & 1.36 \\
96 & 51.5 & 60.9 & 55 & 1.36 \\
128 & \textbf{51.9} & \textbf{61.2} & 72 & 1.45 \\
\bottomrule
  \end{tabular}\vspace{-2mm}
\end{table}

\noindent\textbf{Pre-training of backbone.}
For \cnn, we utilize the 2D convolutional blocks pre-trained on ImageNet for initialization. Here, we compare its performance with the randomly initialized ConvNet. As illustrated in Figure~\ref{fig:pretrain}, leveraging pre-trained models can accelerate the convergence as well as improve the overall performance.

\noindent\textbf{More blocks in early stage.}
We conduct extensive experiments to delve into the scaling rules at different stages.
For our baseline, the block number of four stages is set to [1, 1, 1, 1]. 
Experiments for each stage (see Figure~\ref{fig:num_blocks}) only modifies the number of blocks in the corresponding stage while the block numbers of other stages remain the same.
Figure~\ref{fig:num_blocks}(a) shows the experiments for stage-1, where only the number of blocks in stage-1 is varied, with quantities of 2, 3, 4, and 6, respectively.
Figure~\ref{fig:num_blocks}(b) further ablates the block number of stage-2. For  baseline experiment, the block numbers of the four stages is [2, 1, 1, 1]. the number of blocks in stage-2 is varied with quantities of 2, 3, 4, and 6, respectively. Experiments for stage-3 and stage-4 follows similar practice, setting the block numbers for stages preceding the current one as 2.
The experiments show that increasing the block number of early stage leads to great improvement on performance. While for the later stage, increasing the number of blocks brings almost no performance improvements. 

\noindent\textbf{More depth stage.}
In Section~\ref{sec:bk_design}, we analyze the necessity for adding more depth stages to cover the perception range for large objects. Such analysis is validated by the results presented in Table~\ref{tab:stage5}. It shows that adding one $16\times$ stage, including down-sampling and a ConvNet block, improves the performance by 2.0\% mAP and 1.3\% NDS, while only increasing marginal computational cost.

\noindent\textbf{Removing downsampling in the first stage.} 
As mentioned previously, premature down-sampling may lead to the loss of critical information. To verify the effect of removing downsampling in the early stage, we conduct the experiments and keep the downsampling operation in the first stage (see Table~\ref{tab:stage1_down}). It shows that removing downsampling improves the performance by 7.0\% mAP and 5.4\% NDS, though a large computation cost is introduced.

\noindent\textbf{Input channels of backbone.}
For Pillar-based methods, the input channel quantity of the 2D backbone is equivalent to the number of output channels from the pillar encoder. Table~\ref{tab:ab_input} presents an ablation study to show the effect of varying input channels of 2D backbone on the performance. It indicates that models with larger input channels of the backbone yield much better performance. 
However, increasing the number of input channels introduces much more computational cost. The shown results enable us to scale up the backbone size regarding the number of input channels.

\section{Conclusion}
In this paper, we present \cnn{}, a series of pillar-based 3D object detectors, by exploring 2D backbone scaling and pretraining.
Based on the ConvNext~\cite{ConvNeXt} block, We clearly propose the design rules and scaling principles for the pillar-based 2D backbone.
The marked effectiveness of backbone scaling and pretraining exhibited in this paper provides a promising direction for future works on backbone design. 
The core of our approach revolves around improvements in the backbone. Hence, future modifications on the neck and head parts have the potential to further boost the performance of point cloud 3D detectors. 

\noindent\textbf{Limitation:} Though we present the effectiveness of scaling and pretraining of pillar-based backbone, there are also some limitations for our method. As shown in Table~\ref{tab:stage1_down}, removing downsampling of the early stage greatly improves the performance while introducing much more computation cost. How to deal with high-resolution pseudo-image representation with low computation cost is worth exploring for the near future. Moreover, some generative pretraining strategies, such as masked image modeling~\cite{MAE22, BEiT22}, can be further employed to reconstruct the 2D pseudo image from pillar encoder for high-quality representation, producing good backbone initialization for point cloud 3D object detection.

{\small
\bibliographystyle{ieee_fullname}
\bibliography{egbib}

\begin{thebibliography}{10}\itemsep=-1pt

\bibitem{TransFusion22}
Xuyang Bai, Zeyu Hu, Xinge Zhu, Qingqiu Huang, Yilun Chen, Hongbo Fu, and Chiew{-}Lan Tai.
\newblock Transfusion: Robust lidar-camera fusion for 3d object detection with transformers.
\newblock In {\em CVPR}, 2022.

\bibitem{BEiT22}
Hangbo Bao, Li Dong, Songhao Piao, and Furu Wei.
\newblock Beit: {BERT} pre-training of image transformers.
\newblock In {\em ICLR}, 2022.

\bibitem{nuscenes}
Holger Caesar, Varun Bankiti, Alex~H. Lang, Sourabh Vora, Venice~Erin Liong, Qiang Xu, Anush Krishnan, Yu Pan, Giancarlo Baldan, and Oscar Beijbom.
\newblock nuscenes: {A} multimodal dataset for autonomous driving.
\newblock In {\em CVPR}, 2020.

\bibitem{chen2022milestones}
Long Chen, Yuchen Li, Chao Huang, et~al.
\newblock Milestones in autonomous driving and intelligent vehicles: Survey of surveys.
\newblock {\em IEEE Transactions on Intelligent Vehicles}, 8(2):1046--1056, 2022.

\bibitem{ASPP17}
Liang-Chieh Chen, George Papandreou, Florian Schroff, and Hartwig Adam.
\newblock Rethinking atrous convolution for semantic image segmentation.
\newblock {\em arXiv preprint arXiv:1706.05587}, 2017.

\bibitem{CVCNet20}
Qi Chen, Lin Sun, Ernest Cheung, and Alan~L. Yuille.
\newblock Every view counts: Cross-view consistency in 3d object detection with hybrid-cylindrical-spherical voxelization.
\newblock In Hugo Larochelle, Marc'Aurelio Ranzato, Raia Hadsell, Maria{-}Florina Balcan, and Hsuan{-}Tien Lin, editors, {\em Adv. Neural Inform. Process. Syst. (NeurIPS)}, 2020.

\bibitem{HotSpotNet2020}
Qi Chen, Lin Sun, Zhixin Wang, Kui Jia, and Alan~L. Yuille.
\newblock Object as hotspots: An anchor-free 3d object detection approach via firing of hotspots.
\newblock In Andrea Vedaldi, Horst Bischof, Thomas Brox, and Jan{-}Michael Frahm, editors, {\em ECCV}, 2020.

\bibitem{MVF17}
Xiaozhi Chen, Huimin Ma, Ji Wan, Bo Li, and Tian Xia.
\newblock Multi-view 3d object detection network for autonomous driving.
\newblock In {\em CVPR}, 2017.

\bibitem{largekernel3d2023}
Yukang Chen, Jianhui Liu, Xiangyu Zhang, Xiaojuan Qi, and Jiaya Jia.
\newblock Largekernel3d: Scaling up kernels in 3d sparse cnns.
\newblock In {\em CVPR}, 2023.

\bibitem{VoxelNeXt23}
Yukang Chen, Jianhui Liu, Xiangyu Zhang, Xiaojuan Qi, and Jiaya Jia.
\newblock Voxelnext: Fully sparse voxelnet for 3d object detection and tracking.
\newblock In {\em CVPR}, 2023.

\bibitem{mmdet3d2020}
MMDetection3D Contributors.
\newblock {MMDetection3D: OpenMMLab} next-generation platform for general {3D} object detection.
\newblock \url{https://github.com/open-mmlab/mmdetection3d}, 2020.

\bibitem{ImageNet09}
Jia Deng, Wei Dong, Richard Socher, Li{-}Jia Li, Kai Li, and Li Fei{-}Fei.
\newblock Imagenet: {A} large-scale hierarchical image database.
\newblock In {\em CVPR}, 2009.

\bibitem{VoxelRCNN21}
Jiajun Deng, Shaoshuai Shi, Peiwei Li, Wengang Zhou, Yanyong Zhang, and Houqiang Li.
\newblock Voxel {R-CNN:} towards high performance voxel-based 3d object detection.
\newblock In {\em AAAI}, 2021.

\bibitem{VISTA22}
Shengheng Deng, Zhihao Liang, Lin Sun, and Kui Jia.
\newblock {VISTA:} boosting 3d object detection via dual cross-view spatial attention.
\newblock In {\em CVPR}, 2022.

\bibitem{RepLKNet}
Xiaohan Ding, Xiangyu Zhang, Jungong Han, and Guiguang Ding.
\newblock Scaling up your kernels to 31x31: Revisiting large kernel design in cnns.
\newblock In {\em Proceedings of the IEEE/CVF conference on computer vision and pattern recognition}, pages 11963--11975, 2022.

\bibitem{RepVGG}
Xiaohan Ding, Xiangyu Zhang, Ningning Ma, Jungong Han, Guiguang Ding, and Jian Sun.
\newblock Repvgg: Making vgg-style convnets great again.
\newblock In {\em CVPR}, 2021.

\bibitem{ViT}
Alexey Dosovitskiy, Lucas Beyer, Alexander Kolesnikov, Dirk Weissenborn, Xiaohua Zhai, Thomas Unterthiner, Mostafa Dehghani, Matthias Minderer, Georg Heigold, Sylvain Gelly, Jakob Uszkoreit, and Neil Houlsby.
\newblock An image is worth 16x16 words: Transformers for image recognition at scale.
\newblock In {\em ICLR}, 2021.

\bibitem{SST}
Lue Fan, Ziqi Pang, Tianyuan Zhang, Yu{-}Xiong Wang, Hang Zhao, Feng Wang, Naiyan Wang, and Zhaoxiang Zhang.
\newblock Embracing single stride 3d object detector with sparse transformer.
\newblock In {\em CVPR}, 2022.

\bibitem{FSD22}
Lue Fan, Feng Wang, Naiyan Wang, and Zhao{-}Xiang Zhang.
\newblock Fully sparse 3d object detection.
\newblock In {\em Adv. Neural Inform. Process. Syst. (NeurIPS)}, 2022.

\bibitem{VMVF22}
Hamidreza Fazlali, Yixuan Xu, Yuan Ren, and Bingbing Liu.
\newblock A versatile multi-view framework for lidar-based 3d object detection with guidance from panoptic segmentation.
\newblock In {\em CVPR}, 2022.

\bibitem{MAE22}
Kaiming He, Xinlei Chen, Saining Xie, Yanghao Li, Piotr Doll{\'{a}}r, and Ross~B. Girshick.
\newblock Masked autoencoders are scalable vision learners.
\newblock In {\em CVPR}, 2022.

\bibitem{ResNet16}
Kaiming He, Xiangyu Zhang, Shaoqing Ren, and Jian Sun.
\newblock Deep residual learning for image recognition.
\newblock In {\em CVPR}, 2016.

\bibitem{AFDetV2}
Yihan Hu, Zhuangzhuang Ding, Runzhou Ge, Wenxin Shao, Li Huang, Kun Li, and Qiang Liu.
\newblock Afdetv2: Rethinking the necessity of the second stage for object detection from point clouds.
\newblock In {\em AAAI}, 2022.

\bibitem{PointPillars}
Alex~H. Lang, Sourabh Vora, Holger Caesar, Lubing Zhou, Jiong Yang, and Oscar Beijbom.
\newblock Pointpillars: Fast encoders for object detection from point clouds.
\newblock In {\em CVPR}, pages 12697--12705, 2019.

\bibitem{PillarNeXt}
Jinyu Li, Chenxu Luo, and Xiaodong Yang.
\newblock Pillarnext: Rethinking network designs for 3d object detection in lidar point clouds.
\newblock In {\em CVPR}, pages 17567--17576. {IEEE}, 2023.

\bibitem{UVTR2022}
Yanwei Li, Yilun Chen, Xiaojuan Qi, Zeming Li, Jian Sun, and Jiaya Jia.
\newblock Unifying voxel-based representation with transformer for 3d object detection.
\newblock In {\em Adv. Neural Inform. Process. Syst. (NeurIPS)}, 2022.

\bibitem{BEVMAE22}
Zhiwei Lin and Yongtao Wang.
\newblock Bev-mae: Bird's eye view masked autoencoders for outdoor point cloud pre-training.
\newblock {\em arXiv preprint arXiv:2212.05758}, 2022.

\bibitem{Swin}
Ze Liu, Yutong Lin, Yue Cao, Han Hu, Yixuan Wei, Zheng Zhang, Stephen Lin, and Baining Guo.
\newblock Swin transformer: Hierarchical vision transformer using shifted windows.
\newblock In {\em ICCV}, 2021.

\bibitem{ConvNeXt}
Zhuang Liu, Hanzi Mao, Chao{-}Yuan Wu, Christoph Feichtenhofer, Trevor Darrell, and Saining Xie.
\newblock A convnet for the 2020s.
\newblock In {\em CVPR}, 2022.

\bibitem{Link23}
Tao Lu, Xiang Ding, Haisong Liu, Gangshan Wu, and Limin Wang.
\newblock Link: Linear kernel for lidar-based 3d perception.
\newblock In {\em CVPR}. {IEEE}, 2023.

\bibitem{ERF}
Wenjie Luo, Yujia Li, Raquel Urtasun, and Richard~S. Zemel.
\newblock Understanding the effective receptive field in deep convolutional neural networks.
\newblock In Daniel~D. Lee, Masashi Sugiyama, Ulrike von Luxburg, Isabelle Guyon, and Roman Garnett, editors, {\em NeurIPS}, 2016.

\bibitem{VoxelMAE22}
Chen Min, Dawei Zhao, Liang Xiao, Yiming Nie, and Bin Dai.
\newblock Voxel-mae: Masked autoencoders for pre-training large-scale point clouds.
\newblock {\em arXiv preprint arXiv:2206.09900}, 2022.

\bibitem{HVPR21}
Jongyoun Noh, Sanghoon Lee, and Bumsub Ham.
\newblock {HVPR:} hybrid voxel-point representation for single-stage 3d object detection.
\newblock In {\em CVPR}, 2021.

\bibitem{Pointformer21}
Xuran Pan, Zhuofan Xia, Shiji Song, Li~Erran Li, and Gao Huang.
\newblock 3d object detection with pointformer.
\newblock In {\em CVPR}, 2021.

\bibitem{PillarNet}
Guangsheng Shi, Ruifeng Li, and Chao Ma.
\newblock Pillarnet: Real-time and high-performance pillar-based 3d object detection.
\newblock In Shai Avidan, Gabriel~J. Brostow, Moustapha Ciss{\'{e}}, Giovanni~Maria Farinella, and Tal Hassner, editors, {\em ECCV}, Lecture Notes in Computer Science, 2022.

\bibitem{VGG15}
Karen Simonyan and Andrew Zisserman.
\newblock Very deep convolutional networks for large-scale image recognition.
\newblock In Yoshua Bengio and Yann LeCun, editors, {\em ICLR}, 2015.

\bibitem{EfficientNet19}
Mingxing Tan and Quoc~V. Le.
\newblock Efficientnet: Rethinking model scaling for convolutional neural networks.
\newblock In Kamalika Chaudhuri and Ruslan Salakhutdinov, editors, {\em Proc. Int. Conf. Mach. Learn. (ICML)}, 2019.

\bibitem{FCOSLiDAR22}
Zhi Tian, Xiangxiang Chu, Xiaoming Wang, Xiaolin Wei, and Chunhua Shen.
\newblock Fully convolutional one-stage 3d object detection on lidar range images.
\newblock In {\em Adv. Neural Inform. Process. Syst. (NeurIPS)}, 2022.

\bibitem{PointAugmenting21}
Chunwei Wang, Chao Ma, Ming Zhu, and Xiaokang Yang.
\newblock Pointaugmenting: Cross-modal augmentation for 3d object detection.
\newblock In {\em CVPR}, 2021.

\bibitem{wang2022performance}
Ke Wang, Tianqiang Zhou, Xingcan Li, and Fan Ren.
\newblock Performance and challenges of 3d object detection methods in complex scenes for autonomous driving.
\newblock {\em IEEE Transactions on Intelligent Vehicles}, 8(2):1699--1716, 2022.

\bibitem{wang2023multi}
Li Wang, Xinyu Zhang, Ziying Song, et~al.
\newblock Multi-modal 3d object detection in autonomous driving: A survey and taxonomy.
\newblock {\em IEEE Transactions on Intelligent Vehicles}, 8(7):3781--3798, 2023.

\bibitem{P2P22}
Ziyi Wang, Xumin Yu, Yongming Rao, Jie Zhou, and Jiwen Lu.
\newblock {P2P:} tuning pre-trained image models for point cloud analysis with point-to-pixel prompting.
\newblock In {\em NeurIPS}, 2022.

\bibitem{Argoversev2}
Benjamin Wilson, William Qi, Tanmay Agarwal, John Lambert, Jagjeet Singh, Siddhesh Khandelwal, Bowen Pan, Ratnesh Kumar, Andrew Hartnett, Jhony~Kaesemodel Pontes, Deva Ramanan, Peter Carr, and James Hays.
\newblock Argoverse 2: Next generation datasets for self-driving perception and forecasting.
\newblock In Joaquin Vanschoren and Sai{-}Kit Yeung, editors, {\em Proceedings of the Neural Information Processing Systems Track on Datasets and Benchmarks 1, NeurIPS Datasets and Benchmarks 2021, December 2021, virtual}, 2021.

\bibitem{SECOND}
Yan Yan, Yuxing Mao, and Bo Li.
\newblock {SECOND:} sparsely embedded convolutional detection.
\newblock {\em Sensors}, 2018.

\bibitem{3DSSD20}
Zetong Yang, Yanan Sun, Shu Liu, and Jiaya Jia.
\newblock 3dssd: Point-based 3d single stage object detector.
\newblock In {\em CVPR}, 2020.

\bibitem{HVNet20}
Maosheng Ye, Shuangjie Xu, and Tongyi Cao.
\newblock Hvnet: Hybrid voxel network for lidar based 3d object detection.
\newblock In {\em CVPR}, 2020.

\bibitem{3DVID20}
Junbo Yin, Jianbing Shen, Chenye Guan, Dingfu Zhou, and Ruigang Yang.
\newblock Lidar-based online 3d video object detection with graph-based message passing and spatiotemporal transformer attention.
\newblock In {\em CVPR}, 2020.

\bibitem{centerpoint}
Tianwei Yin, Xingyi Zhou, and Philipp Krahenbuhl.
\newblock Center-based 3d object detection and tracking.
\newblock In {\em CVPR}, pages 11784--11793, 2021.

\bibitem{cia-ssd}
Wu Zheng, Weiliang Tang, Sijin Chen, Li Jiang, and Chi-Wing Fu.
\newblock Cia-ssd: Confident iou-aware single-stage object detector from point cloud.
\newblock {\em arXiv preprint arXiv:2012.03015}, 2020.

\bibitem{Cylinder3D20}
Hui Zhou, Xinge Zhu, Xiao Song, Yuexin Ma, Zhe Wang, Hongsheng Li, and Dahua Lin.
\newblock Cylinder3d: An effective 3d framework for driving-scene lidar semantic segmentation.
\newblock {\em arXiv preprint arXiv:2008.01550}, 2020.

\bibitem{FastPillars}
Sifan Zhou, Zhi Tian, Xiangxiang Chu, Xinyu Zhang, Bo Zhang, Xiaobo Lu, Chengjian Feng, Zequn Jie, Patrick~Yin Chiang, and Lin Ma.
\newblock Fastpillars: A deployment-friendly pillar-based 3d detector.
\newblock {\em arXiv preprint arXiv:2302.02367}, 2023.

\bibitem{VoxelNet}
Yin Zhou and Oncel Tuzel.
\newblock Voxelnet: End-to-end learning for point cloud based 3d object detection.
\newblock In {\em CVPR}, 2018.

\bibitem{CBGS19}
Benjin Zhu, Zhengkai Jiang, Xiangxin Zhou, Zeming Li, and Gang Yu.
\newblock Class-balanced grouping and sampling for point cloud 3d object detection.
\newblock {\em arXiv preprint arXiv:1908.09492}, 2019.

\end{thebibliography}
}

\newpage

\vfill

\end{document}